\documentclass[11pt,a4paper]{article}
\usepackage[hyperref]{emnlp2020}
\usepackage{times}
\usepackage{color}
\usepackage{latexsym}
\usepackage{algorithm}
\usepackage{bm}

\usepackage{colortbl}
\definecolor{LightCyan}{rgb}{0.88,1,1}
\newcolumntype{g}{>{\columncolor{LightCyan}}c}

\definecolor{Orange}{rgb}{1,0.7,0}
\newcolumntype{o}{>{\columncolor{Orange}}c}

\definecolor{Blou}{rgb}{0,0.7,1}
\newcolumntype{f}{>{\columncolor{Blou}}c}

\usepackage{microtype}

\aclfinalcopy 

\usepackage{microtype} 

\usepackage{amssymb}   
\usepackage{amsmath}   
\usepackage{amsfonts}  
\usepackage{amsthm}    
\usepackage{latexsym}  
\usepackage{stmaryrd}  
\usepackage{graphicx}  
\usepackage{color}     
\usepackage{xcolor}    

\usepackage{url}       
\usepackage{xspace}    
\usepackage{etoolbox}  
\usepackage{hyperref}  

\usepackage{booktabs}  
\usepackage{array}     
\usepackage{multirow}  
\usepackage{multicol}  
\usepackage{hhline}    
\usepackage{dcolumn}   
\usepackage{makecell}
\usepackage{epigraph}

\usepackage{tikz}      
\usepackage{tikz-qtree}
\usepackage{pgfplots}  
\pgfplotsset{compat=1.14}

\usepackage{algorithm}
\usepackage{algorithmic}

\usepackage{subcaption}

\usepackage{footnote}
\usepackage{footmisc}
\renewcommand{\thefootnote}{\arabic{footnote}}

\newcommand{\astfootnote}[1]{
\let\oldthefootnote=\thefootnote
\setcounter{footnote}{0}
\renewcommand{\thefootnote}{\fnsymbol{footnote}}
\footnote{#1}
\let\thefootnote=\oldthefootnote
}

\newcommand{\ourname}{\textsc{ProbUNK}}

\title{Extracting the Unknown from  Long Math Problems}

\author{Ndapa Nakashole \\
        Computer Science and Engineering \\
        University of California, San Diego \\
        La Jolla, CA 92093 \\
        \texttt{\{nnakashole\}@eng.ucsd.edu}}
\date{}

\begin{document}
\maketitle
\begin{abstract}
 
In problem solving, understanding the problem that one
seeks to solve is an essential initial step.
In this paper, we propose computational  methods for facilitating problem understanding through the  task of   recognizing the unknown in  specifications of long Math problems. We focus on the topic of Probability.  Our experimental results  show that learning models yield strong results on the task --- a promising first step towards human interpretable, modular  approaches to understanding long Math problems.

\end{abstract}

\section{Introduction}
\textit{``It is foolish to answer a question that you do not understand. It is sad to work for an end that you do not desire}{\rm --- George Polya \cite{george1957solve}}.
According to Mathematician Polya's famous problem solving heuristics, asking general common sense questions about the problem  facilitates problem understanding. One such question is  ``What is the unknown?"~\cite{george1957solve}.  The unknown of a given problem is what the problem requires  to be worked out and solved.  In this paper, we propose computational models for extracting the unknown from  problem specifications.

In online tutorial communities, it is not uncommon to find  problems that include questions like ``what is the solution I  am looking for?"\footnote{\url{https://stats.stackexchange.com/questions/66766/}}. Automated hints such as help with \textit{identifying the unknown}, could  provide timely support to self-directed learners ~\cite{guglielmino1978development,houle1988inquiring,hiemstra1994self}.  Furthermore, information about the unknown can be used as a feature in other tasks that can enhance self-directed learning, such as  similar-question retrieval ~\cite{DBLP:conf/naacl/LeiJBJTMM16}, and question-answer matching~\cite{DBLP:conf/aaai/ShenRJPTX17}, as well as in models for  automated problem solving~\cite{DBLP:conf/nips/SachanDMRX18}.

\begin{figure}[t]
	\centering
	
	\includegraphics[width=\linewidth]{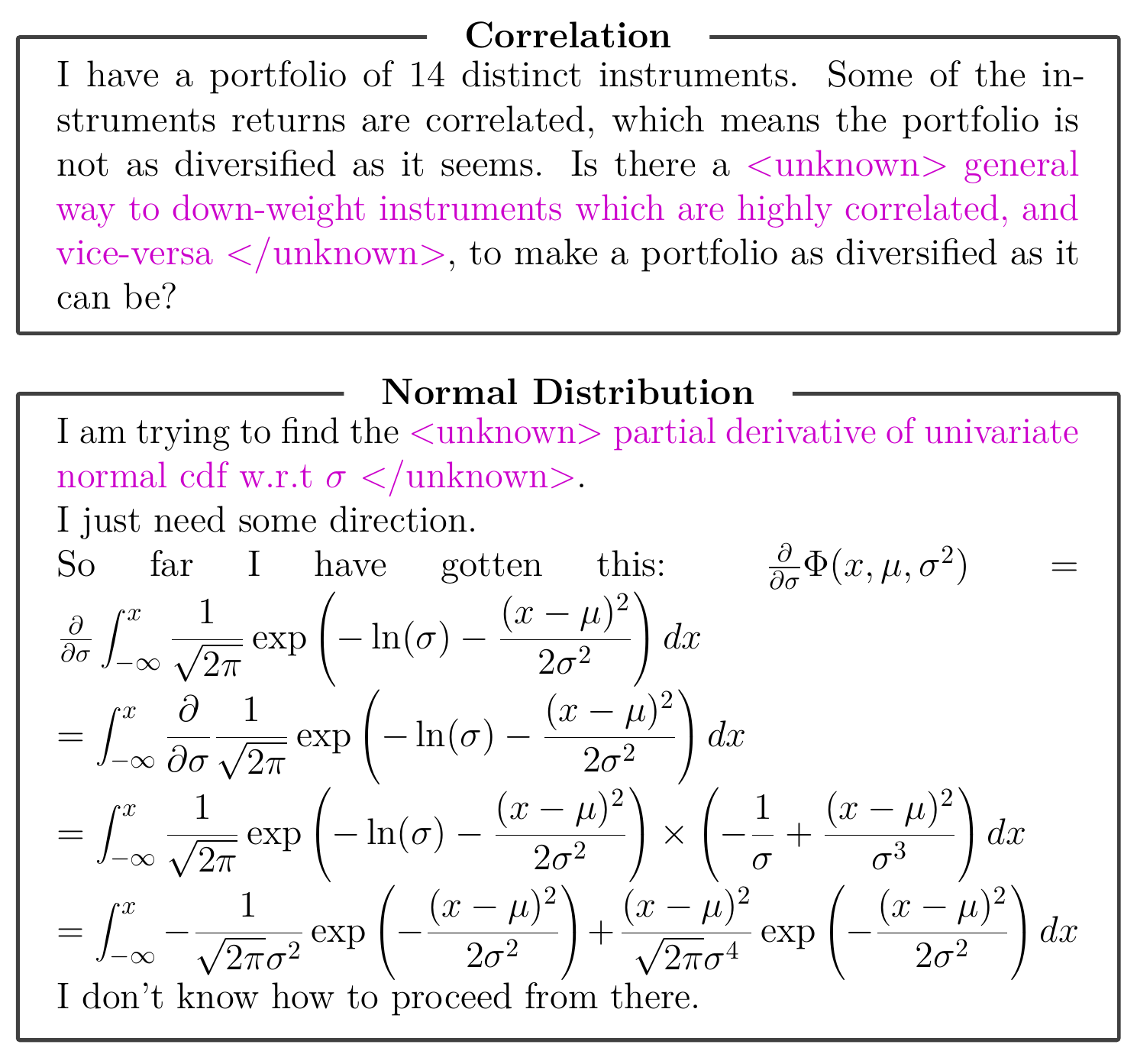}
	\caption{Example problems from \ourname, on the concepts of  \textbf{top:}   Correlation, \textbf{bottom:} Normal Distribution,  labeled with  the unknown.}
	\label{fig:example}
\end{figure}

The unknown is general, it not restricted to problems about a particular  subject-matter.   The problem can be algebraic, geometric, mathematical or nonmathematical~\cite{george1957solve}. 
In this paper, we focus on the topic of \textit{Probability} as a case study.
 To facilitate model training, we created a dataset, \ourname, containing Probability problems labeled with their unknown(s).
 Figure~\ref{fig:example} shows example problems from  \ourname.

\begin{figure*}[t]
	\centering
    \vspace{-4mm}
	\includegraphics[width=0.5\linewidth]{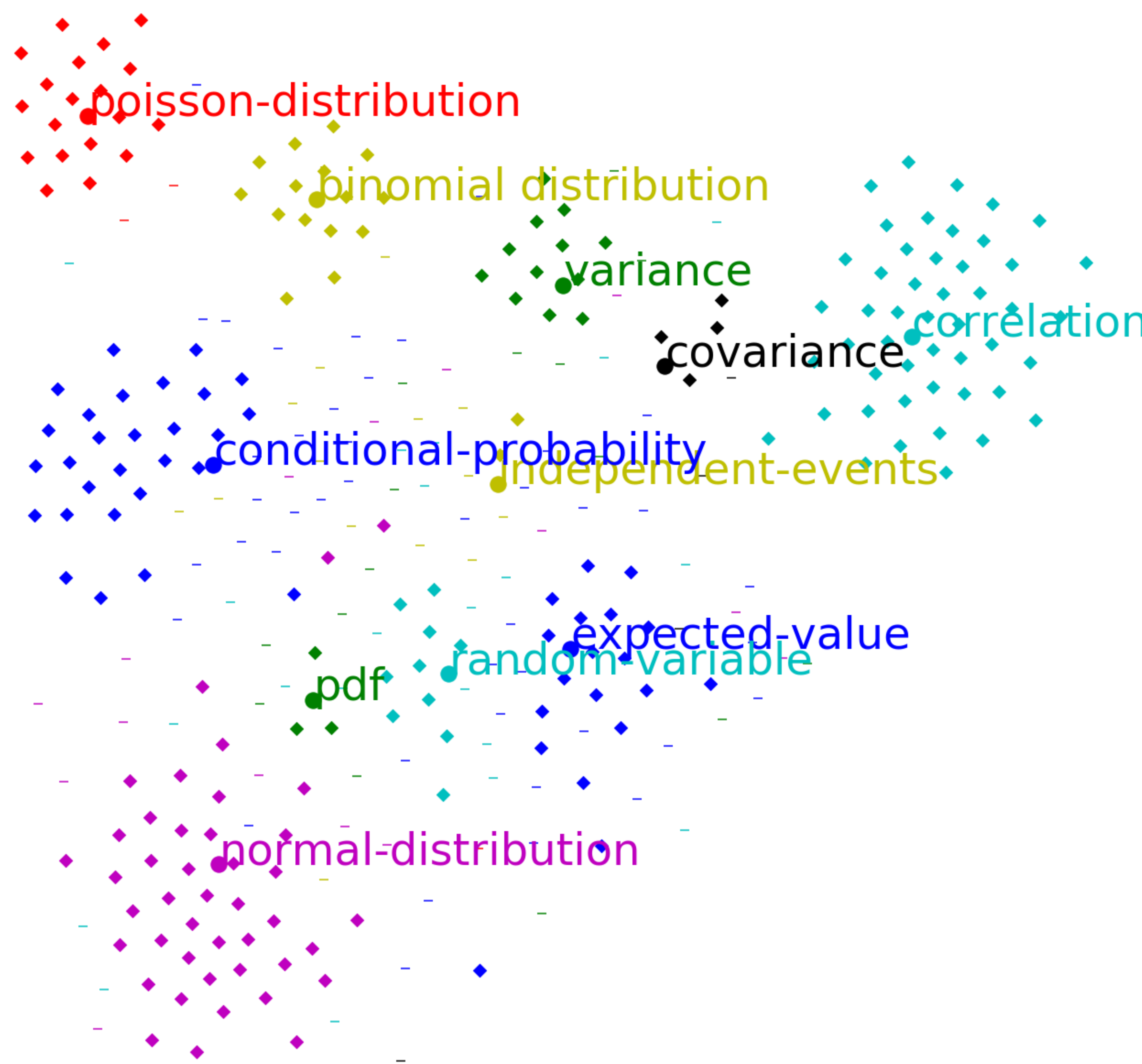}
	\vspace{-2mm}
	\caption{Concept prototypes  (labeled circles)   and prototypical examples (diamonds). }
	\label{fig:prototypes}
\end{figure*}

\paragraph{Prior Work.}
Our work is related to  question understanding, and modular approaches to question answering that advocate decomposing questions into smaller ones ~\cite{DBLP:conf/naacl/AndreasRDK16,DBLP:conf/acl/IyyerYC17,DBLP:conf/emnlp/GuptaL18,DBLP:conf/naacl/TalmorB18,huang2018using,DBLP:conf/acl/GuoZGXLLZ19,DBLP:conf/acl/MinZZH19,DBLP:journals/tacl/WolfsonGGGGDB20}. However, question decomposition in prior work is aimed at enhancing model performance, and is generally not human friendly. Notions of interpretability are provided  through  attention modules~\cite{DBLP:conf/naacl/AndreasRDK16}, or  explicit meaning representations such as first-order logic~\cite{DBLP:conf/nips/SachanDMRX18}, or custom semantic parsing formalisms~\cite{,amini2019mathqa,DBLP:journals/tacl/WolfsonGGGGDB20}. 
In contrast, our goal is to produce output that can be used by  humans, and potentially by  models too.
Additionally,   questions in most prior work are  short, containing just a single sentence. We target longer  multi-sentence problems.



\section{Dataset}
\paragraph{Background Knowledge on Probability.}
As background knowledge, we collected information  about \textit{probability concepts}.  We define a probability concept as a term that is formally defined in the  first five chapters of \citet{wasserman2013all}, and appears in the index at the end of that book. This produced $69$ concepts. For example, `Poisson Distribution',   `Correlation', see  appendix for full list. We augment each concept   with content from  \citet{degroot2012probability}.
On average per concept, we collected $1.6$ definitions  $1.04$ worked-out examples,  and $1.54$ unworked-out examples, from both books.

\begin{table}[t]
\begin{center}
{%
\begin{tabular}{l|c|c|c}
\toprule 
\textbf{} & \multicolumn{1}{|g|}{\textbf{Train}} & \multicolumn{1}{|g|}{\textbf{Dev}} & \multicolumn{1}{|g|}{\textbf{Test}} \\
\hline
{Problems} & {904} & {110}  & {157} \\
{Answers} & {904} & {110}  & {157} \\
 \textbf{Total} & \textbf{1,808} & \textbf{220}  & \textbf{314} \\
\bottomrule
\end{tabular}}
\caption{\label{tab:trainvalidtestsplit} Dataset statistics and split.}
\end{center}
\end{table}

\paragraph{StackExchange Probability Problems.}
We obtained the data dump from \textit{stats.stackexchange},  containing  $281,265$ Statistics problems.  Each problem is labeled with relevant  concept tags. We discarded all    programming-related problems, tagged with: \textit{`Matlab'} and  \textit{`R'}, resulting in  $139, 303$ problems. We also pruned  concept tags  outside of the $69$ background concepts.
To avoid overly complex problems, those with $> 3$ tags were pruned. Lastly, we   only kept problems for which there is an `\textit{accepted answer}' from the forum.  Overall after pre-processing,  there were $1,171$ remaining  problems,  spanning 11 Probability  concepts,  along with their answers.
Table~\ref{tab:trainvalidtestsplit} shows the  train/dev/test  split\footnote{Breakdown of problems by concept is in the appendix.}.

\begin{table}[t]
\begin{center}
 \resizebox{\linewidth}{!}{%
\begin{tabular}{ll|c|c|c}
\toprule 
& & \multicolumn{1}{|g|}{\textbf{Dev}} & \multicolumn{1}{|g|}{\textbf{Test}} &  \multicolumn{1}{|g}{\textbf{Training}}  \\
&\textbf{Method} & \multicolumn{1}{|g|}{\textbf{F1}} & \multicolumn{1}{|g|}{\textbf{F1}}&  \multicolumn{1}{|g}{\textbf{time}}  \\
\hline
1. & MaxEnt &$0.534$  & $0.673$  &04m32s  \\
2. & MLP & $0.547$ & $0.636$  & 06m24s \\
3. & LSTM &  $0.519$ & $0.561$  & 28m41s \\
4. & GRU & $0.537$  & $0.567$  & 25m36s \\
5. & CNN & $\textbf{0.754}$ & $\textbf{0.727}$ & 03m39s \\
\hline

6. &Prototypical Nets(*) & $\textbf{0.787}$  & $\textbf{0.782}$  & 03m21s \\
\bottomrule
\end{tabular}}
\caption{\label{tab:task1classificationresults} Multi-label concept classification. }
\end{center}
\end{table}

\begin{figure*}[t]
	\centering
	\includegraphics[width=\linewidth]{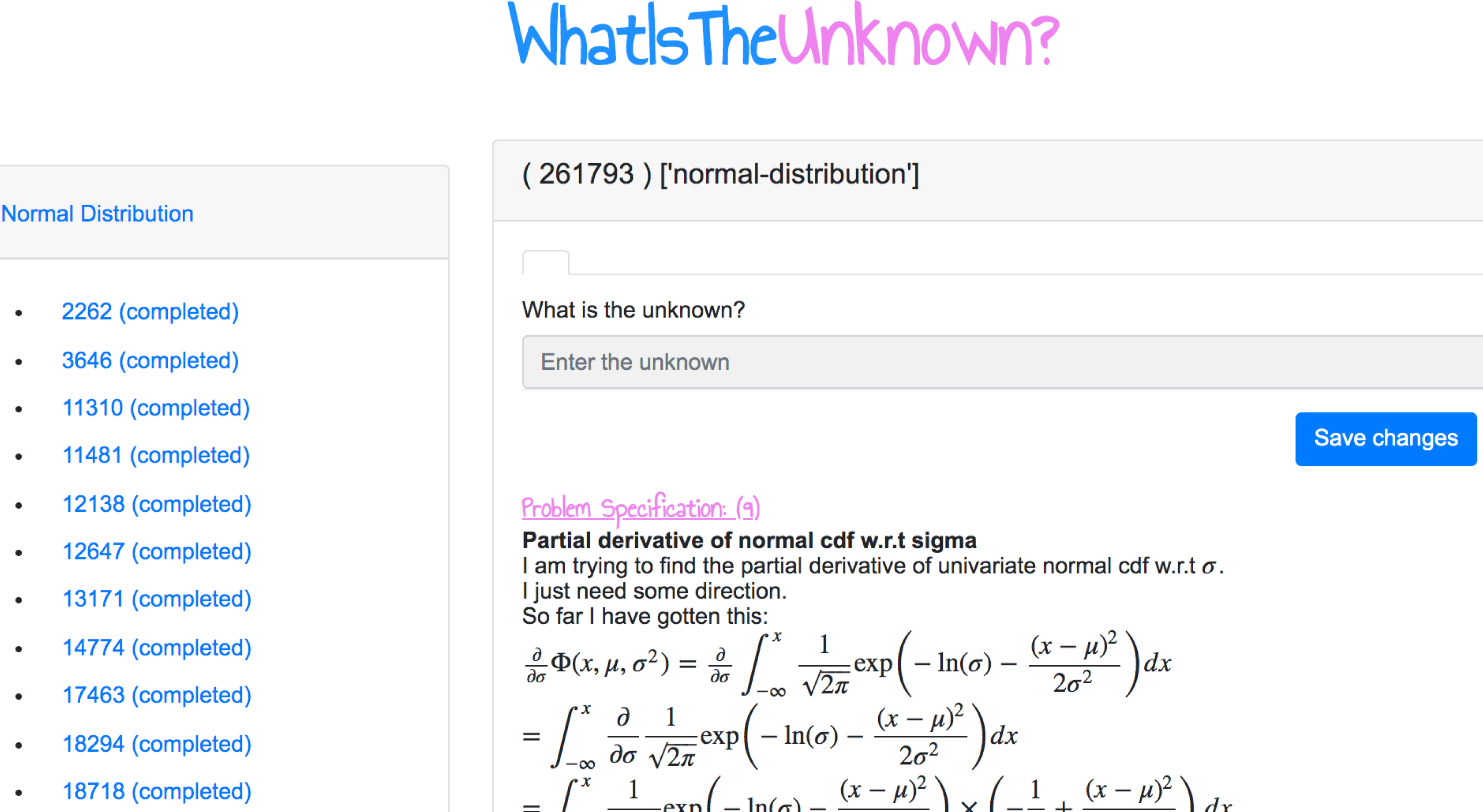}
	\caption{Labeling interface.}
	\label{fig:labelinterface}
\end{figure*}

\paragraph{Concept Classification.}Given the size of our dataset, our setting falls within the limited labeled data regime~\cite{DBLP:conf/nips/PalatucciPHM09,DBLP:conf/nips/SachanDMRX18}. To investigate whether our dataset is sufficient for training models that generalize, we  considered a simple task: classifying problems  by their concepts. This is a multi-label\footnote{$28.9\%$ of the problems in our dataset span $> 1$ concepts.} classification task.  We used a pretrained transformer, BERT~\cite{DBLP:conf/naacl/DevlinCLT19}, to obtain contextualised word embeddings, and then trained\footnote{Training setup details are in the appendix.} various  models.  The results are shown in  Table~\ref{tab:task1classificationresults}. For the  \textbf{MaxEnt} model (a logistic regression classifier), and the  multi-layer perceptron,  \textbf{MLP}, problem representations were obtained by average pooling of the word embeddings. From initial experiments we found that using the CLS token representation, instead of average pooling, degrades performance.
A Convolutional Neural Network (\textbf{CNN}) text classifier, kernel sizes: $3$,$4$,$5$, $6$, each with  $192$ kernels,    has the strongest performance on the full problem,  F1 $\bm {0.73}$ on test data. We also  implemented  \textbf{Prototypical Networks}~\cite{snell2017prototypical}, an approach designed for limited labeled data settings. We used  a CNN (same settings as above)  as the base model for learning prototypical vector representations.
We  evaluated the  Prototypical Network on  problems that are about a single concept, $72.7\%$ of  dev,  and $75.8\%$  of test data, and it produced  strong results, F1~$\bm {0.78}$. Visualizing the learned prototypes and prototypical problems,  Figure \ref{fig:prototypes}  shows that the  learned  representations are meaningful. For instance,  prototypical problems of type \textit{Poisson Distribution} are closest to the prototypical vector of that concept.  This is not trivially expected because there are many prototypical vectors for a concept, which are generated from different support vectors (refer to \citet{snell2017prototypical} for details), and we randomly picked one prototypical vector for each  concept. 

Overall,  both the quantitative results in Table \ref{tab:task1classificationresults}, and the qualitative results in Figure \ref{fig:prototypes},  show that our dataset is adequate for learning
  meaningful  representations,  when using pre-trained contextual embeddings.
\begin{figure}[t]
  \centering
  \begin{subfigure}[b]{0.44\linewidth}
\includegraphics[width=\linewidth]{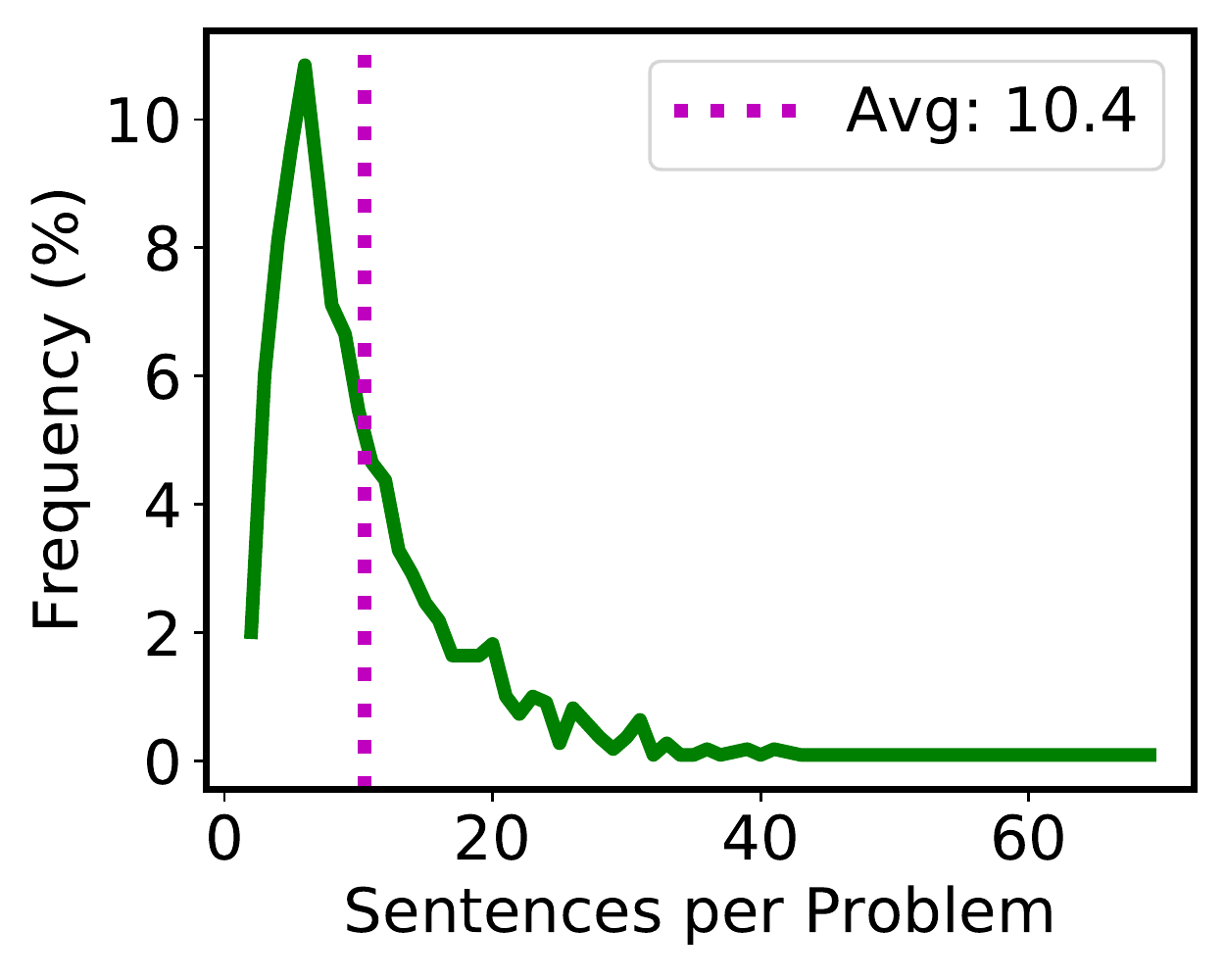}\vspace{-0.25cm}
\caption{}\label{fig:turnspertype}
\end{subfigure}~
  \begin{subfigure}[b]{0.44\linewidth}
\includegraphics[width=\linewidth]{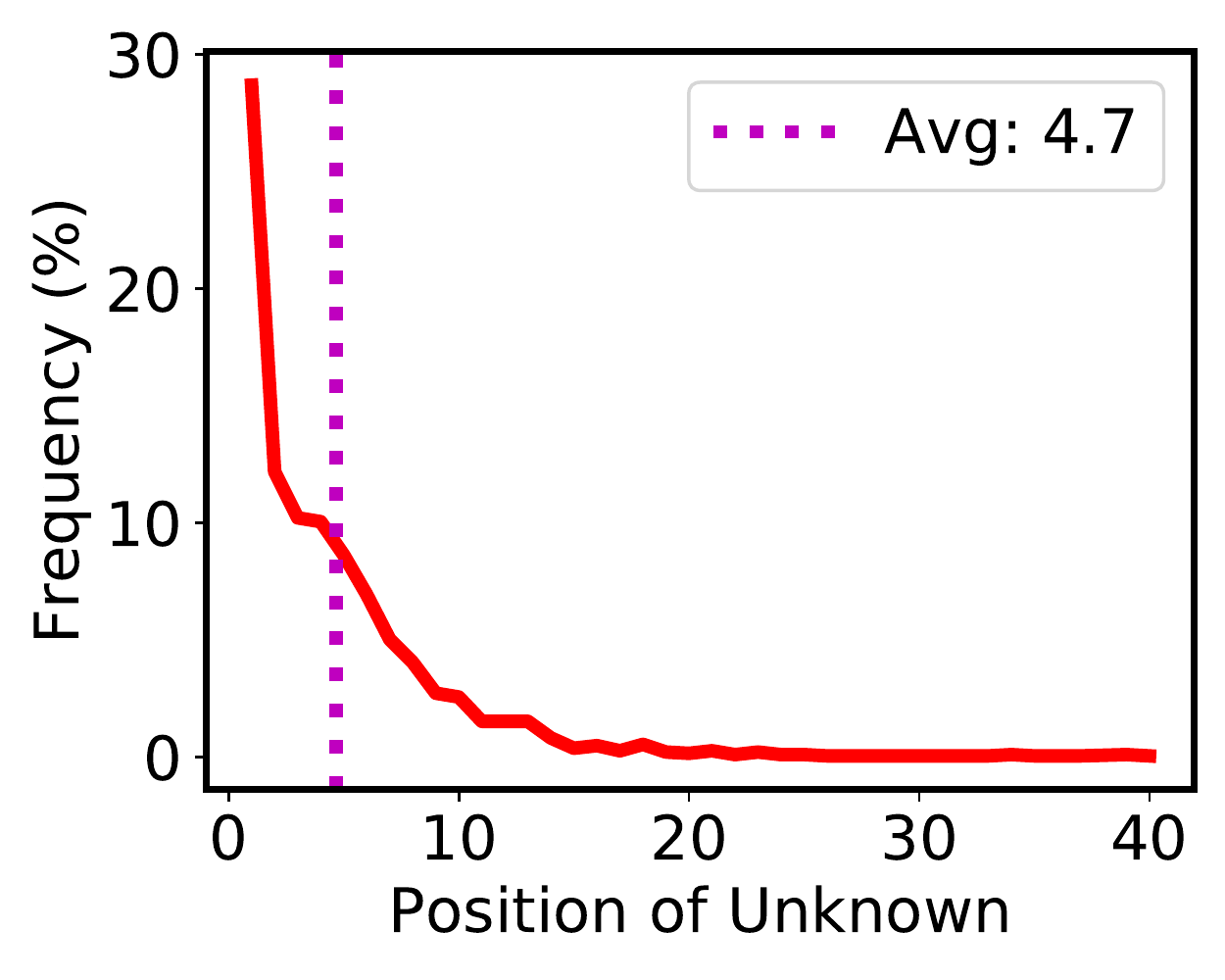}\vspace{-0.25cm}
\caption{}\label{fig:numwords}
\end{subfigure}
\caption{Distributions of  \textbf{(a)} number of sentences per problem, and \textbf{(b)} position of unknown.}
\label{fig:distributions}
\end{figure}

\paragraph{Unknown Annotation.}
To collect labeled data, we  assume  the unknown is  a contiguous sequence of tokens in the problem specification. 
We implemented a simple labeling tool, shown in Figure~\ref{fig:labelinterface}. A problem can have multiple unknowns, each unknown is entered separately.   In total,  $1,171$ problems were annotated. Problems whose unknowns were deemed `unclear'  were few: $26/904(2.9\%)$, $4/110 (3.6)\%$, and $2/157(1.3\%)$ for  train, dev, test, respectively.
On average the unknown(s) is~(are) spread across $1.7$.
 Figures \ref{fig:distributions}(a) and \ref{fig:distributions}(b) show  distributions of sentences per problem,  and  position of the unknown.

\begin{table*}[t]
\begin{center}
\begin{tabular}{ll|c|c|c}
\toprule 
&  & \multicolumn{1}{g|}{\textbf{DEV}} &  \multicolumn{1}{g|}{\textbf{TEST}} & \multicolumn{1}{|g}{\textbf{Training}} \\

& \textbf{Method} & \multicolumn{1}{|g|}{\textbf{F1}}  & \multicolumn{1}{|g|}{\textbf{F1}}  & \multicolumn{1}{|g}{\textbf{time}} \\
\hline
1. & Majority &  $0.456 $ & $0.456 $& n/a\\
2. & $1$st Sentence &  $0.657$ & $0.657 $ & n/a\\
3. & $2$nd Sentence &  $0.499 $ & $0.528 $ & n/a\\
4. & Last Sentence &  $0.531 $& $ 0.508 $ & n/a\\

5. & MaxEnt & $ 0.670$ & $0.686$  & 02m19s  \\

6. & MLP & $ 0.733$ & $0.723$  & 03m06s  \\

7. & CNN &  $0.802 $& $ 0.776 $ & 10m31s\\
8. & CNN\_NoContext &  $0.777 $& $0.759 $ & 09m10s\\
9. & CNN + Graph &  $0.802$ & $\textbf{0.780}$ & 16m42s\\

10. & CNN + Graph + LSTM &  $ 0.504 $ & $ 0.499 $ & 19m48s \\

11. & CNN + Graph + GRU &  $ 0.495 $ & $0.492 $ & 18m44s\\
\bottomrule
\end{tabular}
\caption{\label{tab:task2unknown} Unknown extraction  F1. }
\end{center}
\end{table*}

\section{Approach}

Given a sentence, our goal is to predict if it  contains a sequence of words that describe the unknown, $y_{u}=1$,  or not, $y_{u}=0$. 
Alternative formulation is to treat the problem  a sequence labeling task.  However, in this initial work, the labeled data we obtained is at the sentence level. With the right granularity of labeled data, our task be cast as a sequence labeling task.

\paragraph{Input, Prediction, and Loss.} As input, a problem specification, $Q_i$, and  the $j$-th sentence,  $s_{i,j} \in Q_i $, are presented to a model. 
A context vector, $ {\bm {c}}_i$, is generated from $Q_i$, and  a sentence vector, $\bm{x}_{i,j}$ is generated from $s_{i,j}$.  The two vectors are concatenated to compute  $p_{u}$ as follows:
$$\label{eq:yunk}
    p_{u} = \sigma(\bm{w}^T [\bm{c}_i; \bm{x}_{i,j}] + b),
$$
where  $\sigma$ is the sigmoid function,  $\bm{w}$ and $b$ are learned via the cross entropy  loss  $$ \sum( y_{u} \log p_{u})$$.

\paragraph{Context Vectors.}
To compute the context vector, $\bm{c}_i$, we consider different approaches.

\textbf{\underline{BoW context}}: $\bm{c}_i$ is obtained by average pooling  BERT embeddings of the tokens of $Q_i$ .

\textbf{\underline{CNN Context}}: $\bm{c}_i$ is generated by applying a CNN to $Q_i$.

\textbf{\underline{GCN Context}}:  $\bm{c}_i$ is generated from a Graph Convolutional Network (GCN) ~\cite{DBLP:journals/corr/BrunaZSL13,DBLP:conf/nips/DefferrardBV16,DBLP:conf/iclr/KipfW17}.   
Given a graph with $n$ nodes, the input to a GCN is a feature matrix  $\bm{X} \in \mathbb{R}^{n \times m}$ where a row contains the initial representation,  $x_{u} \in \mathbb{R}^{m}$ of a node $u$. To compute a node's   hidden representation, $h_u \in \mathbb{R}^{d}$, a single graph convolution layer uses the node's neighbors $\bm{N}(u)$,:
$$h_{u} = f\left(\sum_{v \in \bm{N}(u)}\left(\bm{w} x_{v} + b\right)\right) $$,  $f$ is a  non-linear activation function, $\bm{w} \in \mathbb{R}^{d \times m}$ and $b \in \mathbb{R}^{d}$ are learned. We apply the GCN recurrence for $K=3$ steps.
 At  every step, node representations are re-computed  using  the representation of the nodes's neighbors from the previous step, see \citet{DBLP:conf/iclr/KipfW17} for details.

The goal is to use the graph to learn the  kinds of unknowns that pertain to different concepts, as well as to exploit relationships between problems and answers.
We construct a graph 
with three types of \textit{nodes}:  concept, problem;   and answer nodes. We have $5$ \textit{edge} types: \textit{problem-has-type}~(between problems and concepts); \textit{problems-has-answer}~(between problems and answers); and \textit{same-section-as}, \textit{mentioned-in-before-chapters},  \textit{same-chapter-as} (all three between concepts).  We use the concept hierarchy derived from the ordering of concepts in \citet{wasserman2013all} to obtain the last three relationships. Our graph consists of 
$2,360$~nodes, $5,116$ edges,  and $768$  initial node features obtained from  BERT embeddings. Node features are from average pooling the embeddings. For the concept nodes, the  pooled tokens are taken from the concept definitions.
The task for  learning node  representations is predicting the \textit{problem-has-type} links.

\begin{table*}[t]
\centering
\begin{tabular}{l|p{6.5cm}}
\toprule
1. & How could one \textbf{derive} joint pdf of $X, X-k$? \\
2. & How to \textbf{derive} the formula of the t-test of the correlation coefficient \\
\hline
3. & How do you \textbf{calculate} the expected value of $e^{-X}$? \\
4. & I want to \textbf{calculate} $Var\left(\sqrt{\bar{X}/6}\right).$ \\
\hline
5. & \textbf{What is the probability that} the time between trains arriving is 6 minutes or less?\\
6. & \textbf{What is the probability that} $0$ people in the group are fatigued \\
\hline
7. & How do we \textbf{prove that} $(X,Y)$ is not a bivariate normal?\\
8. & I want to \textbf{prove that} the output of $f$ is independent of any input distribution. \\

\bottomrule
\end{tabular}
\caption{Sample positive predictions from dev and test.}
\label{tbl:samples}

\end{table*}
\section{Experimental Evaluation}

 We conducted experiments to evaluate  performance of various models on the task of extracting the unknown.
 All experiments are based on the  train/dev/test split in Table~\ref{tab:trainvalidtestsplit}.  Training  and computing infrastructure details are in the appendix.


    

\paragraph{Methods under Comparison.}Next to each learning method is the number of  parameters.
1)~\textbf{Majority.}~Assigns the most common label, $p_{u}=0$ to every sentence. 
2-4)~\textbf{$\bm n$-th Sentence}~assigns  $p_{u}=1$ to the $\bm n$-th sentence, and  $p_{u}=0$ to all others.
5)~\textbf{MaxEn~($\bm{1,537}$).} Uses BoW for the context vector, $\bm{c}_i$, and  the  sentence vector  $\bm{x}_{i,j}$.
6)~\textbf{MLP~($\bm{1,312,769}$}). Uses the same $\bm{c}_i$ and $\bm{x}_{i,j}$ as  method (5).
7)~\textbf{CNN~($\bm{1,414,275}$).} A~CNN generates both  $\bm{c}_i$, and $\bm{x}_{i,j}$.
8)~\textbf{CNN\_NoContext~($\bm{1,166,467}$).}~Same as (7) but $\bm{c}_i$ is omitted, thus the CNN only encodes the sentence.
9.~\textbf{CNN+Graph~($\bm{1,502,386}$).} Concatenates CNN and  GCN  contexts to get $\bm{c}_i$, whereas  $\bm{x}_{i,j}$ is obtained from just the CNN.
10)~\textbf{CNN+Graph+LSTM~($\bm{2,980,018}$).} $\bm{c}_i$ is as in (9), and $\bm{x}_{i,j}$ is generated by an LSTM.
11)~\textbf{CNN+Graph+GRU~($\bm{2,610,610}$).} Same as (10) but $\bm{x}_{i,j}$ is generated by a GRU.

\paragraph{Results.} Table \ref{tab:task2unknown} shows our main results. 
Among non learning baselines,   the $\bm1$\textbf{st sentence}~(2) baseline  achieves the best performance, F1 is $0.66$ on the test data.
A \textbf{CNN}~(7), well-understood to be a strong text classifier, achieves F1 of $0.776$ on the test data. We see the context vector is important as  \textbf{CNN\_NoContext}~(8) has lower performance than (7).
Incorporating graph structure,  \textbf{CNN+Graph}~(9), yields F1 of $0.78$.  \textbf{ CNN+Graph+LSTM/GRU}~(10-11) both yield poor results. This is likely due to the size of our dataset given that these methods have about twice as many parameters as the other methods, for example, CNN+Graph+LSTM has $2,980,018$ parameters vs. CNN+Graph with $1,502,386$. 

What cues are the methods relying on to achieve strong performance on this task? We sampled some sentences from data, a few of which are shown in Table~\ref{tbl:samples}. It is clear that unambiguous  cues for our task are indeed in the sentences, such as ``derive", ``calculate", ``what is the probability that '', and ``prove that ''. In addition to these sentence cues, our results, (7) vs. (8) and (9) in  Table~\ref{tbl:samples} show that there are also useful signals in the context.

\section{Conclusions and Future Work}
Our new task, dataset, and  results are  only initial steps, and  pave the way for  future work. Beyond  ``what is the unknown?", we can  develop  models that ask other general common sense questions that can support the problem solving process. \citet{george1957solve}  presents a list of  more such questions. These questions are general,  future work includes working on problems from  topics other than Probability. Additionally,  future work can study who this line of work can benefit from well-studied aspects of   knowledge harvesting and representation~\cite{DBLP:conf/webdb/NakasholeTW10,theobald2010urdf,DBLP:conf/vlds/NakasholeSST12, nakashole2012automatic,kumar2017discovering}.


\nocite{*} 
\bibliography{acl2020}
\bibliographystyle{acl_natbib}

   
\section{Dataset}
\subsection{Prototypes Visualization Further Details}
We defined prototypical examples as those whose closest prototypical vector is almost always (in $\geq 95\%$ of the cases) that of the gold truth concept, regardless of the choice of support problems used to generate the prototypical vector.  
We then randomly picked one of the prototypical vectors,  plotted it, and a few of the prototypical example problems.

\subsection{Background Knowledge Concepts}
The 69 Background knowledge concepts collected in our dataset are shown in Table \ref{tab:fullconceptsappendix}. Definitions, along with worked out example for each concept are in the dataset.

\subsection{Stats.StackExchange Preprocessing}
Excluded  and ignored tags are shown in Table \ref{tab:excludedtagsappendix}. If the tag is excluded, we remove from our dataset problems with this tag. If a tag is ignored, we can keep the problem but only if it has other tags besides an ignored tags.

\subsection{Train/Dev/Test by Concept}
Train/Dev/Test split of our dataset listing the number of problems in each partition  per  concept is shown in Table \ref{tab:trainvalidtestsplitappendix}.

\subsection{Concept Classification Training Details}
\paragraph{Implementation.}
In our implementation, we used the Pytorch Library.
\paragraph{Computing Infrastructure.}
OS: 16.04.1-Ubuntu/ x86\_64.  \\
GPU: GeForce GTX 1080 Ti with 12 GB of memory. 

\paragraph{Evaluation Metrics.}
Evaluation metric is Macro f1\_score:
We used the implementation in  sklearn. In python, the code is: \texttt{`from sklearn.metrics import f1\_score}'

\paragraph{Training and Method Details.}
We use the ADAM optimizer, and dropout 0. Word embeddings are initialized with pretrained $768$ dimensional BERT embeddings. The \textbf{MLP} has $3$ hidden layers, and the dimensionality of  hidden states  is $512$. The \textbf{LSTM} is a one layer bidirectional LSTM,  with hidden dimensions  $384$. Concatenating backward and forward outputs of BiLSTM generates  LSTM output vector of size $768$.
The GRU is a one layer bidirectional GRU,  with hidden dimensions of $384$. Concatenating backward and forward outputs of BiGRU generates  GRU output vector of size $768$.

Table \ref{tab:task1classificationresultsappendix} shows performance on the dev set over 11 seeds are used. 
    
Parameter details for the methods for \textbf{concept classification} are in  Table~\ref{tab:task1methods}.\\

\begin{table*}[h]
\begin{center}
 \begin{tabular}{cc}   

\resizebox{0.45\linewidth}{!}{%
\begin{tabular}{l|c|}
\toprule 
\textbf{Concept} & \#. Items \\
\hline
\textbf{Probability} &  \\
1. Probability &  2\\
2. Complement Of An Event &  2\\
3. Disjoint &  3\\
4. Event &  12\\
5. Intersection Of Events &  2\\
6. Monotone Increasing &  2\\
7. Mutually Exclusive &  3\\
8. Partition &  2\\
9. Sample Outcome &  2\\
10. Sample Space &  12\\
11. Set Difference &  1\\
12. Union &  1\\
13. Uniform Probability Distribution &  7\\
14. Independent Events &  11\\
15. Conditional Probability &  13\\
16. Bayes' Theorem &  8\\
\hline
\textbf{Random Variables} &  \\
17. Random Variable &  6\\
18. Cumulative Distribution Function &  6\\
19. Inverse Cumulative Distribution Function &  5\\
20. Probability Density Function &  3\\
21. Probability Function &  4\\
22. Probability Mass Function &  2\\
23. Quantile Function &  5\\
24. The Bernoulli Distribution &  3\\
25. The Binomial Distribution &  9\\
26. The Discrete Uniform Distribution &  3\\
27. The Geometric Distribution &  1\\
28. The Point Mass Distribution &  1\\
29. The Poisson Distribution &  10\\
30. \(t\) Distribution &  1\\
31. Cauchy Distribution &  1\\
32. Exponential Distribution &  2\\
33. Gamma Distribution &  5\\
34. Gaussian &  7\\
35. Normal &  7\\
36. The $\chi^{2}$ &  1\\
\end{tabular}} & 


 \resizebox{0.45\linewidth}{!}{%
\begin{tabular}{l|c|}


37. The Beta Distribution &  4\\
38. The Uniform Distribution &  1\\
39. Bivariate Probability Density Function &  4\\
40. Joint Mass Function &  1\\
41. Marginal Density Function &  4\\
42. Marginal Mass Function &  2\\
43. Independent Random Variables &  5\\
44. Conditional Probability Density Function &  5\\
45. Conditional Probability Mass Function &  3\\
46. Independent And Identically Distributed &  1\\
47. Multinomial &  7\\
48. Multivariate Normal &  1\\
\hline
\textbf{Expectation} &  \\
49. Expected Value &  22\\
50. First Moment &  1\\
51. Mean &  1\\
52. Covariance And Correlation &  2\\
53. Variance &  8\\
54. Conditional Expectation &  9\\
55. Conditional Variance &  2\\
56. Moment Generating Function &  11\\
\hline
\textbf{Inequalities} &  \\
57. Chebyshev’S Inequality &  7\\
58. Hoeffding'S Inequality &  2\\
59. Markov’S Inequality &  3\\
60. Mill'S Inequality &  1\\
61. Cauchy-Schwartz Inequality &  1\\
62. Jensen'S Inequality &  2\\
\hline
\textbf{Convergence of Random Variables} &  \\
63. Converges To $X$ In Distribution &  2\\
64. Converges To $X$ In Probability &  2\\
65. Converges To \(X\) In Quadratic Mean &  1\\
66. The Weak Law Of Large Numbers &  2\\
67. The Central Limit Theorem &  2\\
68. The Delta Method &  1\\
69. The Multivariate Delta Method &  2\\
\bottomrule
\end{tabular}}
\end{tabular}
\caption{\label{tab:fullconceptsappendix}69 Background knowledge concepts. A few concepts are  synonymous, for example 34 and 35.}
\end{center}

\end{table*}

\begin{table*}[h]
\begin{center}
{%
\begin{tabular}{l|c|c|c}
\toprule 
\textbf{Concept} & \textbf{Train} & \#. \textbf{Dev} & \textbf{Test} \\
\hline
The Binomial Distribution & 78 & 8 & 13 \\
Conditional Probability & 135 & 16 & 25 \\
Correlation & 141 & 15 & 26 \\
Covariance & 64 & 8 & 7 \\
Expected Value & 134 & 16 & 21 \\
Independent Events & 73 & 7 & 8 \\
Normal Distribution & 193 & 22 & 29 \\
Probability Density Function & 49 & 4 & 3 \\
The Poisson Distribution & 58 & 6 & 9 \\
Random Variable & 133 & 20 & 26 \\
Variance & 103 & 14 & 17 \\
\hline
\textbf{Total} & \textbf{1,161} & \textbf{136}  & \textbf{184} \\
\hline
\textbf{\#.  Unique Problems} & \textbf{904} & \textbf{110}  & \textbf{157} \\
\textbf{\#. Unique Answers} & \textbf{904} & \textbf{110}  & \textbf{157} \\
 \textbf{Total} & \textbf{1808} & \textbf{220}  & \textbf{314} \\
\bottomrule
\end{tabular}}
\caption{\label{tab:trainvalidtestsplitappendix}Per concept listing of problems in the train/dev/test partitions. Some problems can be categorized under multiple concepts. Each problem has a corresponding accepted answer}
\end{center}
\end{table*}

\begin{table*}[h]
    \centering
    \begin{tabular}{l|c| c}
    \toprule
         \textbf{TAG}& Excluded/Ignored & \textbf{REASON}\\
    \hline
        Matlab &  Excluded &  programming problems\\
        R &  Excluded &  programming problems\\
        Probability& Ignore & Too Generic\\
        Mathematical-statistics& Ignore & Too Generic\\
        Meta-analysis& Ignore & Too Generic\\
        Hypothesis-testing'& Ignore & Too Generic\\
        Distributions& Ignore & Too Generic\\
        Self-study& Ignore & Too Generic\\
        Intuition& Ignore & Too Generic\\
        Definition& Ignore & Too Generic\\
    \bottomrule
    \end{tabular}
    \caption{\textit{stats.stackexchange }data dump pre-processing: excluded  and ignored tags. If the tag is excluded, we remove from our dataset problems with this tag. If a tag is ignored, we can keep the problem but only if it has other tags besides an ignored tags.  }
    \label{tab:excludedtagsappendix}
\end{table*}

\clearpage

\begin{table}[t]
\begin{center}
 {%
\begin{tabular}{ll|c}
\toprule 
& & \multicolumn{1}{|g}{\textbf{Dev}} \\
&\textbf{Method} & \multicolumn{1}{|g}{\textbf{F1}} \\
\hline
1. & MaxEnt &$0.528 \pm 0.003$ \\
2. & MLP & $0.510 \pm 0.020$ \\
3. & LSTM &  $0.474 \pm 0.027$\\
4. & GRU & $0.472 \pm 0.041$ \\
5. & CNN & $\textbf{0.706} \pm 0.028$\\
\end{tabular}}
\caption{\label{tab:task1classificationresultsappendix} Multi-label concept classification dev performance average over all 11 seeds:  [0-5, 10, 100, 1000,10000, 1000000]}
\end{center}
\end{table}
 
\begin{table*}[h]
\begin{center}
{%

\begin{tabular}{l|l | c}
\toprule  
\textbf{MaxEnt} & \#. Parameters & 8,459  \\
 & Epochs (bounds) & [1-300]  \\
 & Epochs (best) &  300  \\
 & Number of epochs search trials & 31 \\
& Choosing hyperparameter values & uniform\ (1, 10, 20, 30, ...\\ 
& Seed & 3 \\
\hline
\textbf{MLP} & \#. Parameters & 924,683  \\
 & Epochs (bounds) & [1-300]  \\
 & Epochs (best) &  300  \\
 & Number of epochs search trials & 31 \\
& Method of choosing epoch values & uniform: 1, 10, 20, 30, ...\\ 
& Seed & 1 \\
\hline
\textbf{LSTM} & \#. Parameters & 4,469,771  \\
 & Epochs (bounds) & [1-300]  \\
 & Epochs (best) &  300  \\
 & Number of epochs search trials & 31 \\
& Method of choosing epoch values & uniform: 1, 10, 20, 30, ...\\ 
& Seed & 2 \\
\hline
\textbf{GRU} & \#. Parameters & 3,583,499  \\
 & Epochs (bounds) & [1-300]  \\
 & Epochs (best) &  300  \\
 & Number of epochs search trials & 31 \\
& Method of choosing epoch values & uniform: 1, 10, 20, 30, ...\\ 
& Seed & 2 \\
\hline

\textbf{CNN} & \#. Parameters & 3,138,829  \\
 & Epochs (bounds) & [1-300]  \\
 & Epochs (best) &  20  \\
 & Number of epochs search trials & 31 \\
& Method of choosing epoch values & uniform: 1, 10, 20, 30, ...\\
& Seed & 10000 \\
& Filter Sizes (bounds) & combinations of [1,2,3,4,5,6]  \\
 & Filter Sizes (best) &  [3,4,5,6]  \\
& Method of choosing Filter Sizes & manual \\

& Filters (bounds) & [50-300]  \\
 & Filters (best) &  [192]  \\
& Method of choosing Filters & manual \\
\hline

\textbf{Prototypical Networks} & \#. Parameters & 2,214,146   \\
&  Episodes (bounds) & [10-200] \\
&  Episodes (best) & 100 \\
& Method of choosing episode values & uniform\\
& Support size &  10 \\
& Query size &  15 \\
\bottomrule
\end{tabular}}
\caption{\label{tab:task1methods}Hyperparameter search for methods in Table 2 of the paper.}
\end{center}
\end{table*}

\clearpage

\section{Experimental Evaluation}

\paragraph{Computing Infrastructure.}
Operating System: 16.04.1-Ubuntu/ x86\_64.  \\
GPU: GeForce GTX 1080 Ti with 12 GB of memory. 

\paragraph{Evaluation Metrics.}
Evaluation metric is Macro f1\_score:
We used the implementation in  sklearn. In python, the code is: \texttt{`from sklearn.metrics import f1\_score}'

\paragraph{Training and Method details}
The \textbf{GCN} input features are  $768$-dimensional, and its hidden states are $100$ dimensional, and the number of convolutional layers is $3$. 

We use the ADAM optimizer, and dropout $0.2$. Word embeddings are initialized with pretrained $768$ dimensional BERT embeddings. The \textbf{MLP} has $3$ hidden layers, and the dimensionalities of  hidden states  is $512$. The \textbf{LSTM} is a one layer bidirectional LSTM,  with hidden dimensions  $384$. 
The \textbf{GRU} is a one layer bidirectional GRU.

Parameter details for the methods for \textbf{unknown extraction} are in  Table~\ref{tab:task2methods}.

\begin{table*}[h]
\begin{center}
{%

\begin{tabular}{l|l | c}
\toprule  
\textbf{MaxEnt} & \#. Parameters & 1,537 
\\
 & Epochs (bounds) & [1-100]  \\
 & Epochs (best) &  30  \\
 & Number of epochs search trials & 11 \\
& Choosing epochs values & uniform\ (1, 10, 20, 30, ...\\ 
& Seed &  3\\
\hline
\textbf{MLP} &  \#. Parameters  & 1,312,769  \\
 & Epochs (bounds) & [1-50]  \\
 & Epochs (best) &  30  \\
 & Number of epochs search trials & 6 \\
& Choosing epochs values & uniform\ (1, 10, 20, 30, ...\\ 
& Seed &  10\\
\hline
\textbf{CNN} &  \#. Parameters  & 1,414,275 \\
 & Epochs (bounds) & [1-100]  \\
 & Epochs (best) &  20  \\
 & Number of epochs search trials & 6 \\
& Choosing epochs values & uniform\ (1, 10, 20, 30, ...\\ 
& Seed &  4\\
& Filter Sizes (bounds) & combinations of [1,2,3,4,5,6]  \\
 & Filter Sizes (best) &  [1,2]  \\
& Method of choosing Filter Sizes & manual \\

& Filters (bounds) & [50-300]  \\
 & Filters (best) &  [192]  \\
& Method of choosing Filters & manual \\
\hline
\textbf{CNN\_NoContext} &  \#. Parameters  & 1,166,467 \\
 & Epochs (bounds) & [1-50]  \\
 & Epochs (best) &  30  \\
 & Number of epochs search trials & 6 \\
 & Seed &  0\\
\hline

\textbf{CNN+Graph} &  \#. Parameters  & 1,502,386 \\
 & Epochs (bounds) & [1-50]  \\
 & Epochs (best) &  30  \\
 & Number of epochs search trials & 6 \\
  & Seed &  0\\
& GCN layers & 3 \\
& GCN hidden state dimensionality &  $100$  \\ 
\hline
\textbf{CNN+Graph+LSTM} &  \#. Parameters  & 2,980,018
 \\
 & Epochs (bounds) & [1-50]  \\
 & Epochs (best) &  30  \\
 & Number of epochs search trials & 6 \\
 & Seed &  10000\\
\hline
\textbf{CNN+Graph+GRU} &  \#. Parameters  & 2,610,610
 \\
 & Epochs (bounds) & [1-50]  \\
 & Epochs (best) &  30  \\
 & Number of epochs search trials & 6 \\
& Seed &  10\\
\bottomrule
\end{tabular}}
\caption{\label{tab:task2methods}Hyperparameter search for methods in Table 3 of the paper.}
\end{center}
\end{table*}

\end{document}